\documentclass[a4paper,conference]{IEEEtran}
\IEEEoverridecommandlockouts
\usepackage{cite}
\usepackage{amsmath,amssymb,amsfonts}
\usepackage{algorithmic}
\usepackage{graphicx}
\usepackage{textcomp}
\usepackage{xcolor}
\usepackage{multirow}
\usepackage{url}
\def\BibTeX{{\rm B\kern-.05em{\sc i\kern-.025em b}\kern-.08em
    T\kern-.1667em\lower.7ex\hbox{E}\kern-.125emX}}

\usepackage[singlespacing]{setspace} 
\newcommand{\rset}{\mathbf{R}}

\newcommand{\G}{\mathcal{G}}
\newcommand{\E}{\mathcal{E}}

\usecounter{algorithmenumi}

\newcommand{\be}{\begin{equation}}
\newcommand{\ee}{\end{equation}}

\newcommand{\bt}{\begin{tabular}}
\newcommand{\et}{\end{tabular}}

\newcommand\pdffig[4][7cm]{
	\begin{figure}[t]
		\centering
		\includegraphics[width=#1]{#2}
		\caption{#3}
		\label{#4}
	\end{figure}
}
\usepackage[right=1.57cm]{geometry}
\begin{document}
%---------------------------------------------------------------------------------------
\title{Unsupervised Abnormal Traffic Detection through Topological Flow Analysis
\thanks{
The authors of this work were supported by a grant of the Ministry of Research, Innovation and Digitization, CNCS/CCCDI - UEFISCDI, project number PN-III-P2-2.1-SOL-2021-0036, within PNCDI III.
Paul Irofti was also supported by a grant of the Romanian Ministry of Education and Research, CNCS - UEFISCDI,
project number PN-III-P1-1.1-PD-2019-0825, within PNCDI III.
Andrei P\u atra\c scu was also supported by a grant of the Romanian Ministry of Education and Research, CNCS - UEFISCDI, project number PN-III-P1-1.1-PD-2019-1123, within PNCDI III.}

}
%---------------------------------------------------------------------------------------
\author{
\IEEEauthorblockN{%
1\textsuperscript{st} Paul Irofti\IEEEauthorrefmark{1},
2\textsuperscript{nd} Andrei Pătrașcu\IEEEauthorrefmark{2} and
3\textsuperscript{rd} Andrei Iulian Hîji\IEEEauthorrefmark{3}
}
\IEEEauthorblockA{\textit{Research Center for Logic, Optimization and Security (LOS)},
\textit{Department of Computer Science}, \\
\textit{Faculty of Mathematics and Computer Science},
\textit{University of Bucharest},
Bucharest, Romania \\
Contact: \IEEEauthorrefmark{1}\texttt{paul@irofti.net}, 0000-0002-7541-4334,\\
\IEEEauthorrefmark{2}\texttt{andrei.patrascu@fmi.unibuc.ro}, 0000-0002-9293-9386, \\
\IEEEauthorrefmark{3}\texttt{andrei-iulian.hiji@unibuc.ro}, 0000-0003-0959-9227}
}

\maketitle
%---------------------------------------------------------------------------------------
\begin{abstract}
Cyberthreats are a permanent concern in our modern technological world. In the recent years, sophisticated traffic analysis techniques and anomaly detection (AD) algorithms have been employed to face the more and more subversive adversarial attacks. A malicious intrusion, defined as an invasive action intending to illegally exploit private resources, manifests through unusual data traffic and/or abnormal connectivity pattern.
Despite the plethora of statistical or signature-based detectors currently provided in the literature, the topological connectivity component of a malicious flow is less exploited. Furthermore, a great proportion of the existing statistical intrusion detectors are based on supervised learning, that relies on labeled data. By viewing network flows as weighted directed interactions between a pair of nodes, in this paper we present a simple method that facilitate the use of connectivity graph features in unsupervised anomaly detection algorithms. We test our methodology on  real network traffic datasets and observe several improvements over standard AD.
\end{abstract}


%---------------------------------------------------------------------------------------
\section{Introduction}

\noindent Nowadays computer security has become a necessity brought by the fast evolution of information technologies. 
The expansion of the network architectures, such as cloud computing, revealed increasingly higher number of threats than before. According to 2021 Cyberthreat Defense Report \cite{Cyber}, the percentage of organizations compromised 
by successful attacks rose by 5.5 \%, which seems to be the largest in the last 7 years. These attacks include malware, ransomware, Denial-of-Service (DoS) and Advanced Persistent Threats (APT).
Despite the fact that security research investments are on a positive trend, alleviation of these threats is not optimistic \cite{Cyber}. However, the constant progress and performance of the recent Intrusion Detection Systems (IDS) bring a  positive light on the subject.

One side of the modern IDSs approaches the intrusion detection as an anomaly detection problem. Detecting outliers in finite samples of data is an old statistical topic, alongside with the design of robust estimators that are resistant to corrupted data points \cite{Hub:04}. From this viewpoint, detecting a network intrusion reduces to learning a statistical estimator that is capable to distinguish between normal and abnormal traffic. However, there are two obvious issues to address. Related to first, one could find a plethora of results in the literature that confirm the efficiency of this statistical approach for usual attacks such as DoS, Probe, User to Root, Remote to User etc. However, in large networks, there is often the case when the attacker uses a stolen set of credentials to obtain data (or access) from (to) multiple nodes of the networks. In this case, while the traffic parameters may seem close to normal, the change in the graph connectivity pattern could reveal an abnormal behavior. In his malicious pursuit, the attacker will probably walk from node to node, through lateral movement, on paths that a normal user would never follow. Therefore, the underlying graph representation of the network traffic, where the nodes are the computers and the edges are the traffic sessions, becomes necessary. Secondly, the real network traffic is by nature not labeled.  Although the most challenging, unsupervised AD methods seems the most intuitive approach of an intrusion detection task.  

In this paper we bring a preliminary evidence showing that the graph connectivity may be an important asset in some cases for unsupervised intrusion detection. We design a simple processing strategy of given flow data that enrich the feature vectors with additional graph embeddings. Our graph embedding method is based on computing egonets of each network node and extract their key features. After training on the extended features, the accuracy of several unsupervised AD algorithms shows slight improvements.

%---------------------------------------------------------------------------------------

\subsection{Related work}
Several wide-range anomaly detection techniques that are often used in network IDSs are listed as follows: Statistical Profiling with Histograms \cite{Ilg:95,Yam:01}, Parametric and Non-parametric Statistical Modeling \cite{Gwa:05,Yeu:02}, (Deep or Shallow) Artificial Neural Networks and Autoencoders \cite{Gam:20,Cho:21,Ahm:21,Yin:17}, (One-Class) Support Vector Machines \cite{Gor:09,Per:06}, Reconstruction methods \cite{Vas:16,Shy:03,Kir:20}, Clustering methods \cite{Ahm:18,Yao:18}. However, generally most of these learning systems detect abnormal data flows or packets based on their features and characteristics. Besides the track imprinted in these features, many attacks manifest their tracks into anomalous underlying connection graph, and therefore graph anomaly detection techniques become an important tool for more insight \cite{Ako:15}.

Graph embedding is used in \cite{He:12}, where the flow data is viewed as an entropy time series, whose features are mapped as nodes in an undirected graph. Here, after computing weights on edges based on covariance between features, the authors devise an algorithm that assign an anomaly score on each flow. Spectral decomposition methods are applied in  \cite{Che:16} to intrusion detection problem. Their method keeps only statistical and spectral features of a given connectivity graph to detect traffic anomalies. In \cite{Cap:19} are used attack graphs to analyze the state evolution of multi-layered attacks in a vulnerable system. We mention that the vertices in these graphs are the attack states and actions, since they serve to modeling of the causality of vulnerability exploitation.

In \cite{Xia:20} the authors devise an IDS that, based on a double graph embedding, expand an original set of features into a new one containing graph embedding information. Their overall approach is vaguely similar to ours, however the embedding procedure and classification algorithms are not related. In the final, they used supervised learning algorithms to classify enhanced features of datasets CIDDS-001 and CIC-IDS2017. 
\vspace{5pt}

\noindent \textit{Paper structure}. In the following Section we describe our graph embedding and feature expansion procedures. We evaluate the empirical performance of these embedding procedures, in Section~\ref{sec:experiments}, by comparison with the application of traditional anomaly detectors onto several well-known datasets. Lastly, we discuss and interpret our result in Section~\ref{sec:results}.

%---------------------------------------------------------------------------------------

\section{Methodology}
\label{sec:methodology}

\noindent As presented in the introduction, the main steps of our method reduce to: $(i)$ embedding of the network flows into a directed graph; $(ii)$ extraction of several statistical node features from the graph and expand the original feature set. We use notation $X \in \rset^{m \times N}$ for flow data, where $m$ is the number of features of a given flow and $N$ the number of flows. 

\vspace{5pt}

\noindent First, given a set of fixed IPs within a network, mapping them  into integers set $[n]:=\{1, 2, \cdots, n\}$, where $n$ is the number of machines in the network, is straightforward. Now we further consider the graph $\G = \{V,E,W\}$, where the set of vertices $V = \{1,2, \cdots, n\}$, $E$ is the set of edges between nodes, corresponding to connections between pairs of IPs, and $W$ is a weight matrix. For instance, given a flow representation between two IPs let $(i,j) \equiv (\text{source}\_IP,\text{destination}\_IP)$, then  $(i,j) \in E$ if there exists a flow between IPs mappings $(i,j)$ and the value $w_{ij}$ on $i$th column and $j$th line in matrix $W$ defines some summable feature, for example the number of packets transmitted between source and destination. 

\begin{figure}[t]
    \centering
    \includegraphics[width=7cm,height=5cm]{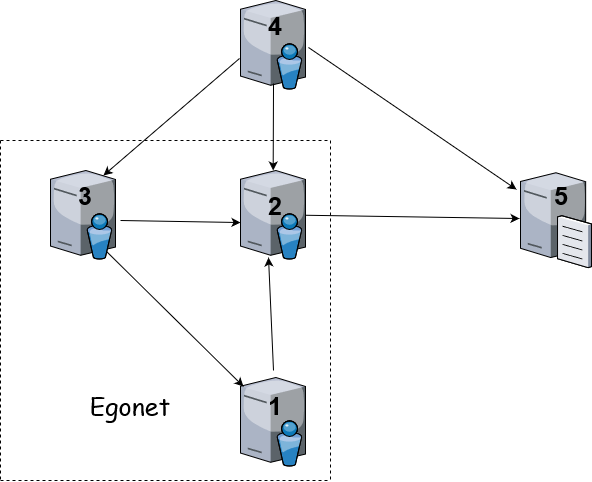}
    \caption{Subgraph enclosed by dashed line represents the egonet of node 1.}
    \label{fig:egonet}
\end{figure}

An \textit{egonet} of node $i$ is defined as the subgraph formed by all neighbors linked to node $i$ \cite{Ako:15}, as described by Figure \ref{fig:egonet}. Notice that egonets associated to different nodes may have different dimensions, depending on the degree of each node. 

\noindent Mainly, our scheme consists of the following three steps:

\vspace{5pt}

\noindent I. \textbf{Flow-to-graph}. The first step performs the conversion of data from flow format into graph format, by retaining source, destination addresses $(i,j)$ and a particular attribute which represents the weight $w_{ij}$. This particular attribute may be any real-valued summable feature in the original data $X$. Since multiple flows may occur multiple times between the same pair of nodes, we get multiple weights $w_{ij}^t$, where $t$ is time counter. We sum over $t$ these weights in order to obtain a final weight: $w_{ij} = \sum_t w_{ij}^t$.

Based on the obtained graph features and weights, we form the directed graph associated with our data. 

\vspace{5pt}

\noindent II. \textbf{Graph-to-features}.
Now on this resulted graph we perform the following operations:
\begin{enumerate}
    \item[1)] Extract all the egonets and stack them into $\E$, where each $\E_i \in \E$ is the egonet associated to node $i \in V$.
    \item[1')] Extract a  random-walk of size $\ell$  for each node. Denote $\E_i \in \E$ as the random-walk associated to node $i$ in $V$. Starting at node $i$, for at most $\ell$ iterations, a neighbor of the current node is randomly chosen  (w.r.t. a uniform probability distribution) and its associated edge is added to the subset $\E_i$. The new chosen node becomes the current node and a new iteration is performed. If either the node $i$ or the walk length $\ell$ are reached, the process terminates and outputs the walk.
    \item[2)] For any $i \in [n]$, extract $p$ features of the egonet/random walk instance $\E_i$. Denote $z_i \in \rset^p$ the vector of these features. 
    \item[3)] Output matrix $Z \in \rset^{p \times n}$, as the array containing all egonet features.
\end{enumerate}
First we perfom only once a single step of the two alternatives $1)$ or $1')$. Notice that the random-walk $\E_i$ computed in scenario $1')$ is not limited to the egonet neighborhood
of node $i$. The statistical features computed in step $2)$, after step $1)$, include: dimension of egonet, the number of out-links, the number of in-links. In alternative scenario $1')$ they include the weight on the first leg of the walk or the weight transferred all the way from the first node to the last one of the walk. The full description of all features can be find in \cite{graphomaly}.

\vspace{5pt}

\noindent III. \textbf{Feature expansion}. Lastly, we expand the original data by adding the columns of $Z$ as prolongation of columns in $X$. Thus, for a given flow $x_t \in X$ from source $i$ to destination $j$, we form: 
$$\hat{x}_t = \begin{bmatrix}x_t \\ z_i \\ z_j \end{bmatrix} \in \rset^{m+2p}.$$ 
The matrix $\hat{X}$ containing columns $\hat{x}_t$ for $t \in [N]$ is the output of our scheme.

\noindent First, notice that the graph embedding at step II maps the flows from $X $ with size $m \times N$ into a final matrix $Z$ with size $p \times n$. By comparison, a column sample of $X$ corresponds to an edge/flow in the graph, while in $Z$ a column associates with a node. In the next section, we show the performance of AD tools in detecting  anomalous nodes.    

Second, the step III is equivalent with inserting local topological information into flow features. Therefore the attacks that forces anomalous connections between machines are likely to be reflected into graph features $\{z_i,z_j\}$ and detected by an usual anomaly detection. 

We further test the performance of several anomaly detections such as: One Class-SVM, Isolation Forests and Local Outlier Factor, onto the data output of the above processing procedure. 

%---------------------------------------------------------------------------------------
%------------
\begin{table*}[t]

\tabcolsep 5pt

\caption{Maximum balanced accuracy and running times when tuning parameters on 1\% of the available data.}
\label{table:grid-search-1p}
\small
\begin{center}
\bt{c c l | c r | c r | c r | c r | c r | c }
Dataset&Method&($m$,$N$,outliers) &
\multicolumn{2}{c|}{OC-SVM} &
\multicolumn{2}{c|}{LOF} &
\multicolumn{2}{c|}{IForest} &
\multicolumn{1}{c}{Ensemble} \\
\hline
%----------------------------------------------
\multirow{3}{4em}{CIC-IDS2017}
& standard & (87, 4588, 5)
& 0.8751 & 0.81s
& 0.9751 & 0.01s
& 0.8751 & 0.07s
& \textbf{0.9152}   \\
%----------------------------------------------
& graph & (48, 382, 1)
& \textbf{0.9252} & 0.02s
& 0.9685 & 0.01s
& 0.4738 & 0.04s
& 0.8972  \\
%----------------------------------------------
& mixed & (183, 4588, 5)
& 0.8167 & 5.55s
& \textbf{0.9755} & 2.23s
& \textbf{0.9010} & 0.07s
& 0.8698  \\
%----------------------------------------------
\hline
%----------------------------------------------
\multirow{3}{4em}{UNSW-NB15}
&standard&(59, 4400, 321)
& 0.5831 & 2.43s
& \textbf{0.8663} & 0.46s
& \textbf{0.9584} & 0.70s 
& \textbf{0.9511}   \\
%----------------------------------------------
&graph&(48, 42, 4)
& \textbf{0.7829} & 0.01s
& 0.7763 & 0.01s
&0.6053 & 0.03s 
& 0.5   \\
%----------------------------------------------
&mixed&(155, 4400, 321)
& 0.5831 & 3.81s
& 0.8037 &9.73s
& 0.9216 & 0.14s
& 0.9123   \\
%----------------------------------------------
\et
\end{center}
\end{table*}
%----------------------------------------------------------
\begin{table*}[t]

\tabcolsep 5pt

\caption{Balanced accuracy and running times
when training on 10\% of the data
from the UNSW-NB15 and CIC-IDS2017 datasets
with the parameters obtained in Table~\ref{table:grid-search-1p}
}
\label{tab:cic2017-unsw-training-10p}
\small
\begin{center}
\bt{c c l | c c | c c | c c | c c | c c | c c}
Dataset & Method &($m$,$N$,outliers) &
\multicolumn{2}{c|}{OC-SVM} &
\multicolumn{2}{c|}{LOF} &
\multicolumn{2}{c|}{IForest} &
\multicolumn{1}{c}{Ensemble} \\
\hline
%----------------------------------------------
\multirow{3}{4em}{CIC-IDS2017}
& standard &(87, 45883, 928)
& 0.3724 & 242.30s
& \textbf{0.4811} & 529.12s
& 0.4132 & 0.35s 
&  \textbf{0.4996}  \\

& graph & (48, 1999, 1)
& 0.4997   & 0.43s
& 0.4705 & 0.07s
& \textbf{0.4750} & 0.06s
& 0.4874  \\

& mixed &(183, 45883, 928)
& \textbf{0.5955} & 601.00s
& 0.4805 & 86.42s
& 0.4222 & 0.32s
& 0.4821   \\
%----------------------------------------------
\hline
%----------------------------------------------
\multirow{3}{4em}{UNSW-NB15}
& standard&(59, 44004, 5148)
& 0.6542 & 358.21s
& 0.5096 & 147.76s
& 0.7259 & 5.09s 
& 0.5474  \\

& graph&(48, 46, 4)
& \textbf{0.7829} & 0.01s
& \textbf{0.6382} & 0.01s
& 0.3289 & 0.03s
& 0.5119  \\

& mixed & (155, 44004, 5148)
& 0.6619 & 579.19s
& 0.5775 & 750.70s
& \textbf{0.7926} & 0.82s
& \textbf{0.9103}   \\
%----------------------------------------------
\et
\end{center}
\end{table*}
%------------

%---------------------------------------------------------------------------------------

\section{Experiments}
\label{sec:experiments}

In this section we are interested in seeing numerical results of enhancing data
with graph specific features.
In our simulations we use
One-Class SVM~(OC-SVM)~\cite{Gor:09,Per:06},
Local Outlier Factor~(LOF)~\cite{Bre00},
Isolation-Forest~(IForest)~\cite{Liu:08},
and an ensemble~\cite{ensembles} that includes the above.
In the implementation of the latter we use voting methods~\cite{vote}.
In our tables and figures,
"standard" denotes the results on the plain data from the public datasets,
"graph" denotes the results on the data aggregated in the form of a graph,
and "mixed" the results on the plain data with the added graph features.

Even though we focus here on shallow machine learning methods,
which we prefer for their performance, speedy results, and known theoretical properties,
we also performed preliminary tests with autonecoder and variatonal autoencoder architectures
that have not yet shown promising results on the experimental setup presented here.

The experiments were performed using public datasets.
% ISCX-IDS-2012 (\url{https://www.unb.ca/cic/datasets/ids.html}), 
% a dataset created by the  Canadian Institute of Cyber Security 
% using specific profiles for generating traffic for
% HTTP, SMTP, SSH, IMAP, POP3 and FTP \cite{Cic2012}.
It contains different features of the flows which have been generated using the IBM QRadar appliance.
CIC-IDS2017\footnote{\url{https://www.unb.ca/cic/datasets/ids-2017.html}}
%is created by the same institution by
is created by the Canadian Institute of Cyber Security 
simulating the benign behavior of 25 users
and replicating a series of attacks.
Flows extracted from this traffic contain 85 features.
UNSW-NB15 \footnote{\url{https://research.unsw.edu.au/projects/unsw-nb15-dataset}} is a dataset
which contains 49 features for the flows extracted using Bro-IDS and Argus tools.
Here, IXIA PerfectStorm tool was used to generate the underlying network traffic.

In order to run the experiments a dedicated station with 32 AMD Ryzen Threadripper PRO 3955WX CPUs was used.
Our implementation relied on the following software packages, among others:
pyod 0.9.6 \cite{pyod},
scikit-learn 1.0.2,
tensorflow 2.7.0,
graphomaly 0.1.

In line with our methodology,
we assume that we have access to a small initial dataset
depicting the normal state of the computers nodes inside the network
through their recorded traffic layer-3 traffic.
Thus,
in our experiments,
we only extract the first 1\% of samples from each dataset
and assume that this is known data with known labels on which we can initially train our models.
Even though we are only interested in the unsupervised setting,
the labels help us tune the parameters through grid-search techniques.
The datasets are laid out as time series,
meaning that the first selected samples reflect exactly the scenario described above.
We denote with $m$ the number of features and with $N$ the number of samples.

% The chosen public datasets contained multiple large files,
% so we decided to use just a single file from each dataset in our test scenario.
% For the ISCX-IDS-2012 dataset,
% we split the data in 10\% train data and 90\% test data.
% For the other 2 datasets, because of the substantially larger number of features,
% we chose to use 1\% as training data and 99\% as testing data.

% We designed 3 different scenarios in order to show different approaches for the selection of the features used to train our models. In the first scenario(\ref{table:1})

%---------------------------------------------------------------------------------------

\begin{table}[t]

\tabcolsep 5pt

\caption{Types of Attacks detected on the UNSW-NB15 dataset with the ensembles from Table \ref{tab:cic2017-unsw-training-10p} (does not apply to the graph method)}
\label{tab:unsw-attacks}
\small
\begin{center}
\bt{l |l |r |r }
Dataset & 
Attack &
Detected &
Total \\
\hline
%----------------------------------------------
\multirow{9}{4em}{standard}
& Exploits
& 163
& 2088  \\

& DoS
& 79
& 1014  \\

& Fuzzers
& 29
& 516  \\

& Worms
& 0
& 7  \\

& Backdoor
& 11
& 138  \\

& Analysis
& 9
& 123  \\

& Shellcode
& 2
& 52  \\

& Reconnaissance
& 31
& 548  \\

& Generic
& 256
& 662  \\
\hline

%----------------------------------------------
\multirow{9}{4em}{mixed}
& Exploits
& 1933
& 2088  \\

& DoS
& 911
& 1014  \\

& Fuzzers
& 502
& 516  \\

& Worms
& 7
& 7  \\

& Backdoor
& 124
& 138  \\

& Analysis
& 109
& 123  \\

& Shellcode
& 47
& 52  \\

& Reconnaissance
& 506
& 548  \\

& Generic
& 644
& 662  \\

\et
\end{center}
\end{table}
%------------
%-----------------------------------------------------------
\section{Results}
\label{sec:results}

In Table~\ref{table:grid-search-1p}
we present the grid-search results on 1\% of the data
for both databases when using the standard, graph and mixed features.
The two columns underneath each method represent the balance accuracy~(BA) and the training execution time.
We can see that standard and mixed methods are giving similar BA results,
identical even for OC-SVM with UNSW-NB15,
but the standard ensemble performing better in both cases.
The execution times are lower for the graph methods,
where there is fewer data to process,
and longer for mixed methods where the graph features are added to the standard data.
The experiment objective is to obtain proper parameters to be used in future model training
on data where labels are not available.

Table~\ref{tab:cic2017-unsw-training-10p} uses the parameters obtained in Table~\ref{table:grid-search-1p}
to train the models on the next 10\% of available data from the time-series.
We see a clear degradation in the balanced accuracy compared to the tuned experiments:
the dataset is larger and new attacks are present and the model parameters are not optimal.
For CIC-IDS2017 all three approaches provide
similar results for the methods and the ensemble.
Instead,
on UNSW-NB15 we see an improvement offered by the graph-based approaches.
We assume that this is due to the richer summable attributes in UNSW-NB15
compared to CIC-IDS2017
where most of the attributes are either existing statistics (already summed) or flags information.
In terms of execution times,
we see a proportional increase corresponding to the ten-fold increase in analyzed data-points.

We further investigate the UNSW-NB15 results in Table~\ref{tab:unsw-attacks} where we compare the
standard and mixed ensembles for their capability of identifying specific types of attacks.
By identifying more attack samples,
the mixed method clearly outperforms the standard one in all scenarios.
Worm attacks are not even detected by the standard model.
%---------------
\pdffig[\columnwidth]
    {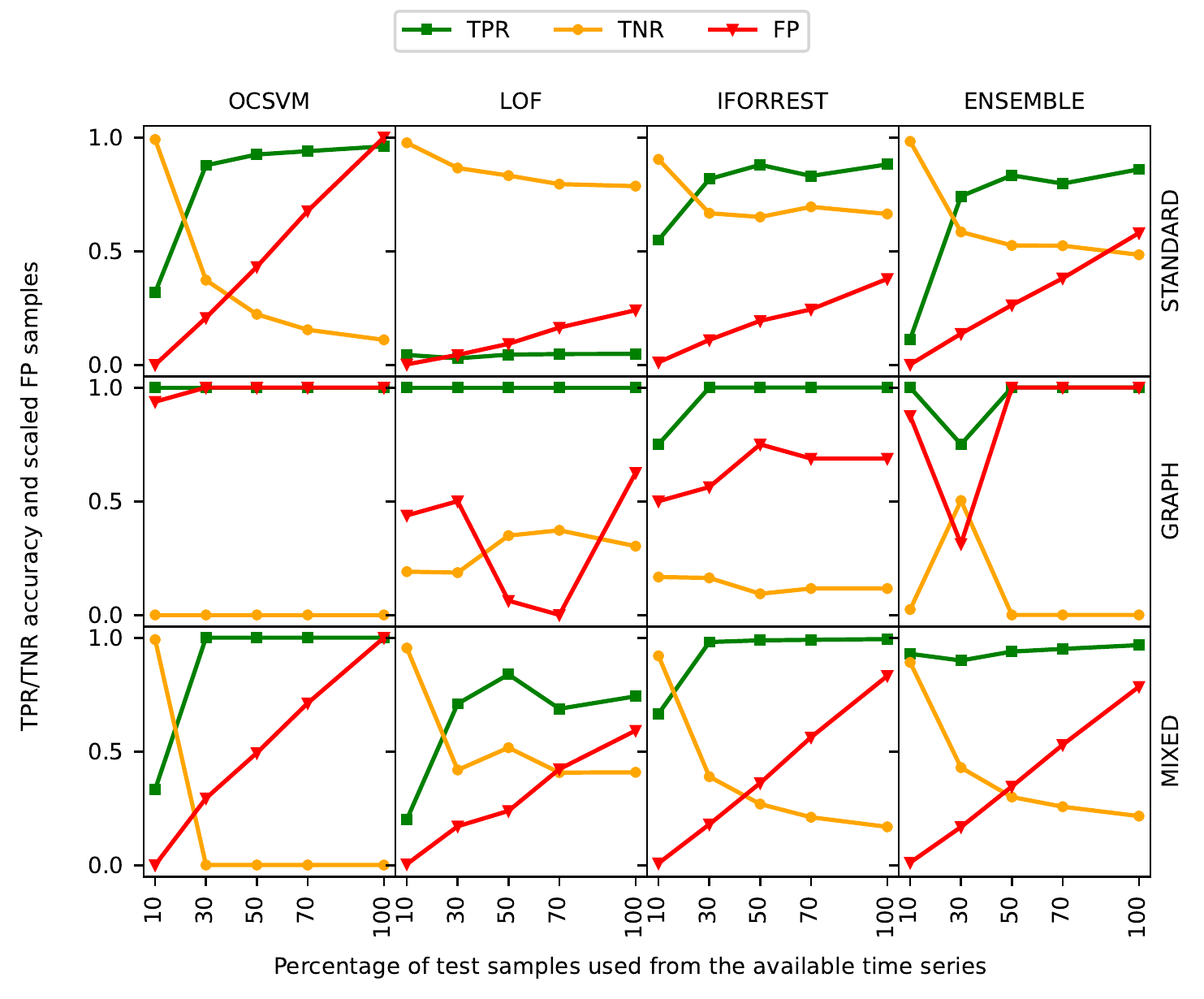}
    {Test results for the models from Table~\ref{table:grid-search-1p} on the UNSW-NB15 dataset.
    For graphical purposes, for each row the number of false positives (FP) are scaled in the $[0,1]$ interval.}
    {fig:unsw}
%---------------
We now use the models form Table~\ref{table:grid-search-1p} as predictors
for the rest of the data samples from the UNSW-NB15 dataset.
Figure~\ref{fig:unsw} depicts the performance for different test dataset sizes:
10\%, 30\%, 50\%, 70\% and 100\%.
The True Positive Rates (TPR)
and
True Negative Rates (TNR) are analyzed
together with the number of False Positives (FP)
for all models, depicted on the columns,
and for all approaches, depicted on the rows.
For each row,
the number of false positives were scaled
such that the reader can see the relative differences
between each method.
As expected,
model performance degrades with time
as normal behaviour evolves and new types of attacks arrive.
We observe that OC-SVM is the most sensible to these changes,
while IForest seems more robust.
Ensembles tend to attenuate false positives and promote good TPR rates.

%--------------------------------------------------------------------------------------
\section{Conclusions}
In this paper we studied the performance of unsupervised machine learning methods
when analyzing computer networks
by starting from a small dataset of known labeled packet samples
that we use to tune model parametrization
which we then use to investigate their performance for further unsupervised learning
on new incoming unlabeled data.
Data is augmented through graph feature extraction techniques,
such as egonets and random walks,
in order to improve the robustness of our models.

%---------------------------------------------------------------------------------------
\bibliographystyle{unsrt}
\bibliography{riss.bib}

\end{document}